\documentclass[11pt]{article}
\usepackage[T1]{fontenc}
\usepackage[utf8]{inputenc}
\usepackage{amsmath,amsfonts,amsthm,amssymb}
\usepackage{setspace}
\usepackage{fullpage}
\usepackage{fancyhdr}
\usepackage{xspace, xcolor}
\usepackage{url}            %

\usepackage{verbatim}
\usepackage{lastpage}
\usepackage{natbib}

\usepackage{titletoc}
\usepackage{booktabs}       %
\usepackage{hyperref}       %
\usepackage{times}
\usepackage{setspace}
\usepackage{graphicx,float,wrapfig}
\usepackage{algorithm}
\usepackage{enumitem}
\usepackage[noend]{algpseudocode}
\usepackage{subcaption}
\usepackage{multirow}
\allowdisplaybreaks

\theoremstyle{definition}

\def\shownotes{1}  %
\ifnum\shownotes=1
\newcommand{\authnote}[2]{[#1: #2]}
\else
\newcommand{\authnote}[2]{}
\fi

\usepackage{amsmath,amsfonts,bm}

\def\1{\bm{1}}

\makeatletter
\newcommand{\ve}{\@ifnextchar\bgroup{\velong}{{\bm{e}}}}
\newcommand{\velong}[1]{{\bm{#1}}}
\makeatother

\DeclareMathAlphabet{\mathsfit}{\encodingdefault}{\sfdefault}{m}{sl}
\SetMathAlphabet{\mathsfit}{bold}{\encodingdefault}{\sfdefault}{bx}{n}

\def\calT{{\mathcal{T}}}

\newcommand{\E}{\mathbb{E}}

\def\({\left(}
\def\){\right)}
\def\[{\left[}
\def\]{\right]}

\newcommand{\ind}[1]{\mathbb{I}\left[ #1 \right]}

\newcommand{\context}{c}
\newcommand{\statement}{s}
\newcommand{\cp}{\rho}
\newcommand{\depth}{\iota}
\newcommand{\useinvoke}{\text{<use\_invoke>}\xspace}
\newcommand{\noinvoke}{\text{<no\_invoke>}\xspace}
\newcommand{\invoke}{\text{<invoke>}\xspace}
\newcommand{\einvoke}{\text{</invoke>}\xspace}
\newcommand{\true}{\text{<true>}\xspace}
\newcommand{\false}{\text{<false>}\xspace}
\newcommand{\unif}{\text{Unif}}
\newcommand{\Null}{\text{Null}}
\newcommand{\Ginvoke}{\widehat{G}}

\newcommand{\method}{\text{ProD}\xspace}

\newcommand{\Ttree}{\calT_\text{tree}}

\setlist[enumerate]{label=(\alph*),itemsep=1pt,topsep=0pt,leftmargin=20pt}

\usepackage{listings}
\lstdefinelanguage{Isabelle}{
	keywords={theorem, lemma, by, if, then, else, <invoke>, </invoke>},
	sensitive=true,
	morecomment=[l]{--},
	morestring=[b]",
	basicstyle=\small\ttfamily, %
	keywordstyle=\bfseries, 
	commentstyle=\color{codegray},
	stringstyle=\color{codegray},
}

\definecolor{lightred}{rgb}{1, 0.95, 0.95}
\definecolor{lightblue}{rgb}{0.95, 0.95, 1}
\definecolor{codegreen}{rgb}{0,0.6,0}
\definecolor{codegray}{rgb}{0.25,0.25,0.25}
\definecolor{codepurple}{rgb}{0.58,0,0.82}
\definecolor{backcolour}{rgb}{0.98,0.98,0.95}

\lstdefinestyle{mystyle}{
	backgroundcolor=\color{lightred},   
	commentstyle=\color{codegreen},
	keywordstyle=\color{magenta},
	numberstyle=\tiny\color{codegray},
	stringstyle=\color{codepurple},
	breakatwhitespace=false,         
	breaklines=true,                 
	captionpos=b,                    
	keepspaces=true,                 
	numbers=left,                    
	numbersep=5pt,                  
	showspaces=false,                
	showstringspaces=false,
	showtabs=false,                  
	tabsize=2,
}
\title{Formal Theorem Proving by Rewarding LLMs to Decompose Proofs Hierarchically}

\author{%
	Kefan Dong\qquad Arvind Mahankali\qquad Tengyu Ma \\\\
	Stanford University \\
	\texttt{\{kefandong,amahank,tengyuma\}@stanford.edu}
}
\date{}

\begin{document}

	\maketitle

	\begin{abstract}
		\noindent Mathematical theorem proving is an important testbed for large language models’ deep and abstract reasoning capability. This paper focuses on improving LLMs’ ability to write proofs in formal languages that permit automated proof verification/evaluation. Most previous results provide human-written lemmas to the theorem prover, which is an arguably oversimplified setting that does not sufficiently test the provers' planning and decomposition capabilities. Instead, we work in a more natural setup where the lemmas that are directly relevant to the theorem are not given to the theorem prover at test time. We design an RL-based training algorithm that encourages the model to decompose a theorem into lemmas, prove the lemmas, and then prove the theorem by using the lemmas. Our reward mechanism is inspired by how mathematicians train themselves: even if a theorem is too challenging to be proved by the current model, a positive reward is still given to the model for any correct and novel lemmas that are proposed and proved in this process. 
		During training, our model proposes and proves lemmas that are not in the training dataset. In fact, these newly-proposed correct lemmas consist of 37.7\% of the training replay buffer when we train on the dataset extracted from Archive of Formal Proofs (AFP). The model trained by our RL algorithm outperforms that trained by supervised finetuning, improving the pass rate from 40.8\% to 45.5\% on AFP test set, and from 36.5\% to 39.5\% on an out-of-distribution test set.
	\end{abstract}

	\section{Introduction}\label{sec:intro}

The reasoning abilities of large language models (LLMs) are a significant marker of artificial intelligence and critical for complex and safety-sensitive applications, e.g., medical diagnosis \citep{singhal2023large,fleming2023medalign}, legal document review \citep{guha2024legalbench},online tutoring \citep{wang2023chatgpt,ruan2024reinforcement}. Yet recent studies highlight the limited performance of LLMs on reasoning tasks (e.g., \citet{mundler2023self,valmeekam2023planning} and references therein). %

\begin{figure}[htp]
	\centering
	\includegraphics[width=\linewidth]{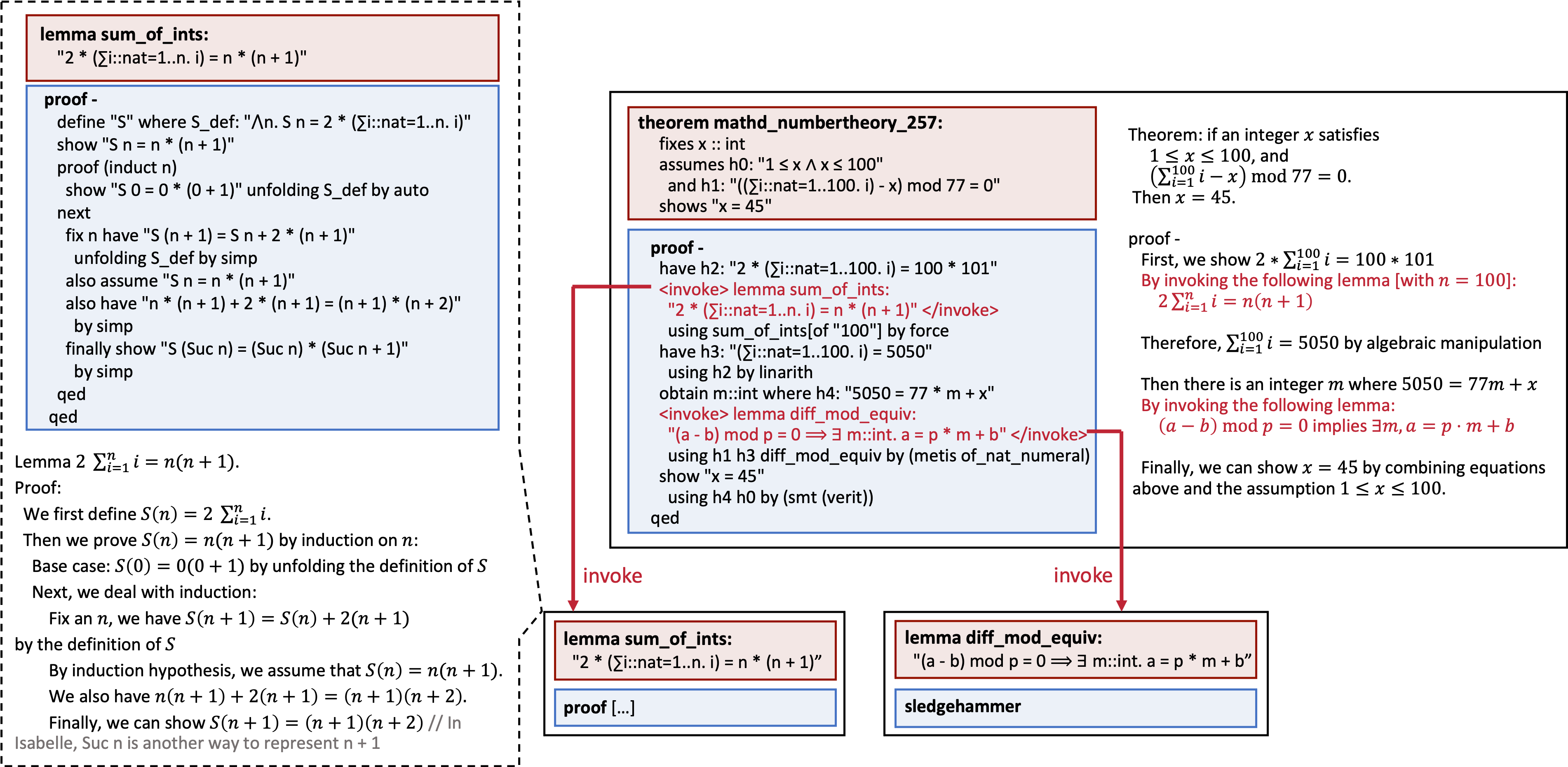}
	\caption{\textbf{Left}: in the dashed callout block, we show an example of an Isabelle proof and its explanation in natural language. \textbf{Right}: an example of a proof tree. The two child nodes correspond to the two new lemmas proposed in the proof of the root node.}
	\label{fig:proof-tree}
\end{figure}
Automated theorem proving by LLMs is an excellent reasoning task that abstracts away the need for numerical manipulation or tool use (e.g., using a calculator) and allows for precise evaluation of correctness with an automatic verifier (such as Isabelle \citep{nipkow2002isabelle} and Lean \citep{de2015lean}) even without ground truth.
Thanks to tools such as  Sledgehammer \citep{paulsson2012three} that can automatically complete low-level details, the granularity of formal proofs is similar to natural language proofs (see Fig.~\ref{fig:proof-tree} (Left) for an illustrative example). 
Note that verifying a proof is fundamentally much easier than generating the proof.\footnote{The former is in P whereas the latter is undecidable in the worst case~\cite{turing1936computable,church1936note}.} Thus, learning to prove theorems from verifiers' supervision is reminiscent of weak-to-strong generalization \citep{burns2023weak}. 

Previous results in this area largely focus on the setting where the theorem prover can use all the lemmas in the formal proof library, including those particularly written to decompose a specific theorem's proof \citep{jiang2021lisa,polu2020generative}.
This setting arguably oversimplifies the problem and doesn’t sufficiently test the models' planning and decomposition capabilities, and it is unclear whether the resulting models can be used to prove new theorems from scratch when such lemmas are not available at test time. Instead, we work in a more natural setup where the theorem prover needs to propose and prove lemmas to decompose the proof hierarchically itself (see Section~\ref{sec:setup} for more details). In Section~\ref{sec:setup-compare}, we demonstrate that this task is much more challenging.

In addition, most existing proof-generation algorithms leverage the formal verifier by (a) providing the verifier's current proof state to the LLMs step-by-step, and (b) using best-first search algorithms such as MCTS to build a multi-step proof from many LLM-generated steps \citep{jiang2022thor,han2021proof}. The major challenge of these methods is the high computation cost incurred because (a) requires re-running LLMs on a different context that consists of a verifier's (long) proof state at every step, and (b) requires generating many proof steps first and then select the best ones. As a concrete example, the search-based algorithm in \citet{lample2022hypertree} requires more than 1K GPU days with A100s to train a model with 600M parameters, whereas our method only takes less than 36 GPU days to train a 7B parameter model.

\begin{figure}[htp]
	\centering
	\includegraphics[width=.80\linewidth]{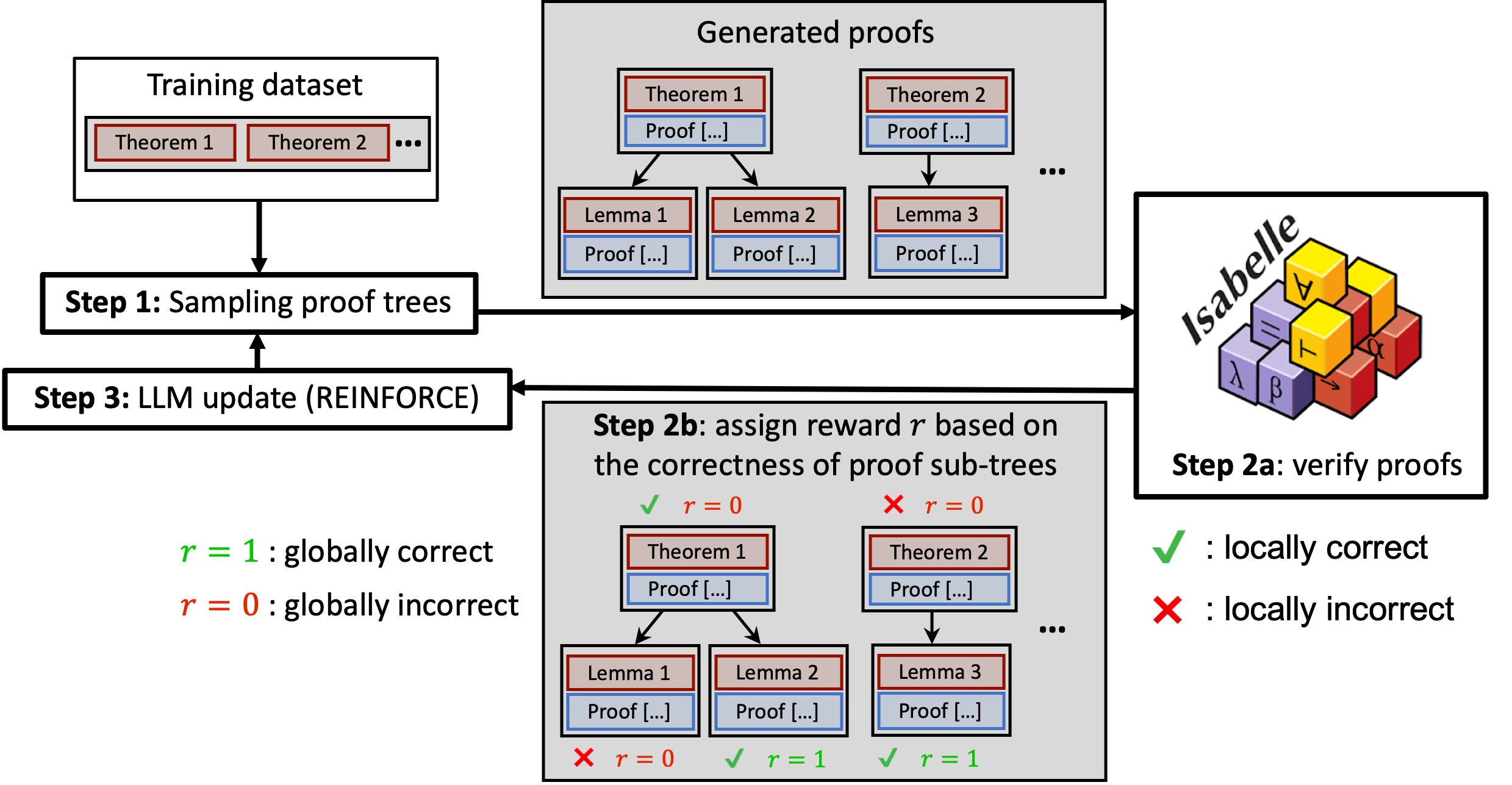}
	\caption{Illustration of our algorithm Proof Decomposer, \emph{\method-RL}. In step 2b, the statement is \textit{locally correct} if it is proved correctly using the proposed lemmas, and it is \textit{globally correct} if all the proposed lemmas are also proved correctly. As an important feature of our algorithm, even if a theorem (Theorem 1) is not proved by the model because some lemmas (Lemmas 1) are not proved, we still train on the correct lemma (Lemma 2) by setting its reward $r=1$.
	}
	\label{fig:alg}
\end{figure}

To address these issues, we design a method, Proof Decomposer (\method), which uses LLMs to hierarchically propose and prove new lemmas and generate complete proofs directly without searching. We augment the formal proofs syntax so that the model can propose new lemmas by including their statements during the proof, and prove these lemmas separately. Hence, a complete proof of a theorem forms a tree structure where the child nodes are the lemmas proposed in the proof of the parent node (Fig.~\ref{fig:proof-tree} (Right)), and the theorem is considered proven only if all the proofs in the tree are correct.

We train our models with reinforcement learning (RL) in a way that somewhat imitates a mathematician’s process: we reward correct partial proofs (i.e., proof sub-trees) even if the original theorem (i.e., the root node) is not proved entirely.
Since our model can generate and prove novel lemmas during training, it can still make progress even if the theorem is too challenging. This is reminiscent of how mathematicians prove standalone lemmas which make progress towards solving an open problem. Our training method can also be viewed as an instantiation of hindsight experience replay \citep{andrychowicz2017hindsight}, where the hindsight trajectories are correct proof sub-trees. We illustrate our algorithm in Fig.~\ref{fig:alg}, and defer the details to Section~\ref{sec:rl}.

We test our model, \method-RL, by generating proof trees on holdout theorems that the model has not been trained on, and show that ProD-RL improves upon existing baselines.
Compared with the supervised fine-tuned (SFT) model, our model improves the pass rate from 40.8\% to 45.5\% on the holdout test set, and from 36.5\% to 39.5\% on out-of-distribution test set, whereas vanilla reinforcement learning without lemma proposals during training does not improve the corresponding SFT model (see Section~\ref{sec:results}).
This is partly because our method encourages the model to propose and prove additional lemmas --- in fact, 37.7\% of the lemmas proved during training are not in the dataset. As a result, the model still improves even if it is already fine-tuned on the same dataset with human-written ground-truth proofs. However, we also observe that our method, \method-RL, is less robust to distribution shifts, and we leave further investigation of this issue as future work.

We release our carefully structured test sets that has a clean separation of the train and test theorems from AFP at \url{https://huggingface.co/kfdong/ProD/tree/main}. We hope that our test bed can help the community develop and validate models that generalize effectively to novel theorems without human-written lemmas.

	\section{Setup}\label{sec:setup}

\paragraph{Conditional proofs.} We use the term \emph{conditional proof} to denote a proof that, in addition to the standard formal proof syntax, can propose new lemmas by enclosing their statements by \invoke and \einvoke tokens (examples shown in the blue boxes of Fig.~\ref{fig:proof-tree}). In particular, a conditional proof has the following format: 
\begin{equation*}
	t_1 \;\invoke\; l_1 \;\einvoke\; t_2 \;\invoke\; l_2 \;\einvoke\; \cdots\; t_k \;\invoke\; l_k \;\einvoke\;  t_{k+1} 
\end{equation*}
where $t_1,\cdots, t_{k+1}$ denote proof segments in the original formal proof syntax (see e.g., Fig.~\ref{fig:proof-tree}, proof texts in black), and $l_1,\cdots,l_t$ denote proposed lemma statements (see e.g. Fig.~\ref{fig:proof-tree}, proof texts in red).\footnote{In this paper, we use the terms `lemma' and `theorem' relatively --- theorem refers to the statement that we are currently focusing on, and lemma refers to the statement proposed during the proof. In other words, there is no fundamental difference between a lemma and a theorem.}

\paragraph{Proof tree nodes.} With the proposed lemmas, a complete proof forms a tree structure (as shown in Fig.~\ref{fig:proof-tree}). 
A node in a proof tree is a tuple of premises, context, a theorem statement, and a conditional proof.
Premises represent the lemmas that are treated as common knowledge, which are typically not directly relevant to the proof. We allow the model to directly use them in the proof so that it does not have to repetitively prove all the fundamental facts, such as properties of continuous functions and natural numbers.
Context represents the necessary contents to prepare the theorem statement, such as the definition of specific objects/functions. We use the context as part of the prompt for the LLMs to generate proofs, and to configure the proof verifier to check the generated proofs.

\paragraph{Correctness of conditional proofs and proof trees.} 
A proof tree node $n$ with conditional proof $$t_1 \;\invoke\; l_1 \;\einvoke\; t_2 \;\invoke\; l_2 \;\einvoke\; \cdots\; t_k \;\invoke\; l_k \;\einvoke\;  t_{k+1}$$ is \emph{locally correct} if, after adding $l_1, …, l_k$ to the set of premises, $t_1 … t_{k+1}$, is a proof to the statement of $n$ that is accepted by the formal verifier under the context of $n$. 

We consider a proof tree valid if, for every node, each of its child nodes corresponds to one proposed lemma and shares the same premises and context with its parent node. A tree node $n$ is \emph{globally correct} with respect to a given set of tree nodes $N$ if we can construct a valid proof tree with root $n$ using the locally correct tree nodes in $N$. We use this more flexible definition of global correctness since if we generate more than one proof tree per theorem, we may mix their locally correct nodes to form a globally correct proof.

Global correctness corresponds to the standard notion of correctness (i.e., whether the theorem is proved), and local correctness is a weaker concept, referring to the correctness of conditional proofs assuming the proposed lemmas. When a tree node is globally correct, we can construct a complete proof to its statement that is acceptable by the formal proof verifier --- first, we build a valid proof tree from the locally correct subset of $N$, and then list all the statements and their corresponding conditional proofs in a child-first order and remove all the lemma proposal steps (since the proposed lemmas and their proofs will be already listed in the proof text according to child-first order).

\paragraph{Dataset construction.} 
We construct the datasets by parsing raw proof-library files into tuples of the form (premises, context, statement, conditional proof). 

In particular, we first segment each of the files into blocks 
$c_1\; s_1\; p_1\; \cdots\; c_l\; s_l\; p_l$
where the $s_i$ are theorem statements, the $p_i$ are the corresponding proofs, and the $c_i$ are the file contents between proofs, such as object definitions and local assumptions. 
Next, we build proof trees from each segmented file by iteratively removing ($s_i$, $p_i$) pairs from the file if the theorem $s_i$ is not referred to in the remaining file contents (in other words, in the first iteration we peel off the root nodes of the proof trees from the file, and then the nodes in the next level, etc.). Note that some theorems cannot be peeled off by this process because they are referred to in some file content $c_j$ (e.g., lemmas used to instantiate local objects). We use $\Ttree$ to denote the subset of theorems peeled off during the process.

For every theorem $s_i$, we construct an example where the context is the concatenation of $\{c_j:j<i\}$ and $\{s_j:j<i,s_j\not\in \Ttree\}$ in the order they appear in the file. That is, we exclude all the lemmas that are ever peeled off --- the remaining lemmas are included in the context.

To construct the conditional proof of theorem $s_i$, we add the proposed lemma statements to the original proof $p_i$. In particular, we split the proof $p_i$ into steps  $t_1,\cdots,t_{k}$ using the formal language parser. Then for every step $t_j$ that uses lemmas $l_{j, 1}, \ldots, l_{j, n_j}$ from $\Ttree$, we insert the statements of these lemmas enclosed by the \invoke and \einvoke tokens, denoted by $\zeta_j= \invoke\; l_{j,1}\;\einvoke\; \cdots\; \invoke\; l_{j,n_j}\;\einvoke$, into the proof right before $t_j$. In other words, the conditional proof is the concatenation $\zeta_1\;t_1\;\cdots\;\zeta_k\;t_{k}$.
Similar to \citet{jiang2022thor}, we use Sledgehammer, a premise selection tool that automatically searches for proofs to the current goal, to replace proof steps that are originally generated by it (see Section~\ref{sec:hyperparameters} for more details) so that the mode can focus less on the tedious low-level details.

The premises are all the theorems from \emph{predecessor} files, which are typically not directly relevant to the theorem (otherwise they will be stated in the same file). Theorems in the premise set can be used directly in the proof, or they can be selected by Sledgehammer to search for proof steps. In our implementation, the premises are implicitly defined by the dependency graphs of the files.

We split the training and test set (AFP test) based on the dependency of the files in the proof library so that the examples in the training set never refer to any files in the test set (see Section~\ref{sec:setup-compare} for details).
We also construct an additional test set, AFP 2023, by parsing AFP files submitted after the knowledge cutoff date of the Llemma model (April 2023) to eliminate potential data leakage issues. 
Compared with prior works \citep{jiang2021lisa,first2023baldur}, the two major differences in our setup are the availability of lemmas from the same file and the training/test split. In Section~\ref{sec:setup-compare}, we discuss and test their effects in detail.

Finally, to construct the SFT dataset, for each example in the training set, if its conditional proof proposes at least one lemma, we create an augmented example by moving the proposed lemmas from the conditional proof into the context --- this augmented example does not propose new lemmas and is always locally correct.
	\section{Methods}\label{sec:methods}
In this section, we first describe how to use LLMs to generate proof trees, and then introduce our reinforcement learning method  (\method-RL) that rewards the model to decompose proofs hierarchically.

\subsection{Generating proof-trees using LLMs}\label{sec:generation}
To generate proof trees using an autoregressive model $\pi_\theta$, we need to first fine-tune the model to follow a specific format:
\begin{enumerate}
	\item the input $x$ to the model $\pi_\theta$ is the concatenation of a context and a theorem statement, and 
	\item the expected output $y$ of the model is a special token $t_0$ followed by a conditional proof, where $t_0$ is either \useinvoke or \noinvoke, denoting whether the following conditional proof should propose new lemmas. 
\end{enumerate}
We introduce the special token $t_0$ before a conditional proof so that we can increase the probability of the \useinvoke token during RL to let the model propose more lemmas for better exploration.

We summarize our proof-tree generation algorithm in Alg.~\ref{alg:tree-generation-test}. Given a theorem statement $s$ and the corresponding context $c$, we first sample from $\pi_\theta$ autoregressively starting with the prompt $x = c\;s$, and ideally the model outputs a special token $t_0$ followed by a conditional proof $\cp$ (Line ~\ref{line:3}). Next, we parse the conditional proof $\cp$ and collect the proposed lemmas $l_1,\cdots, l_k$ for the next round of generation (Line ~\ref{line:5}). We force the model to generate conditional proofs without proposing new lemmas at a certain depth so that the proof tree doesn’t grow indefinitely, which can be implemented easily by replacing the prompt with $x’= c\;s\;\noinvoke$ (Line ~\ref{line:6}).

\begin{algorithm}[htp]
	\caption{Generate proof trees (test time)}
	\label{alg:tree-generation-test}
	\begin{algorithmic}[1]
		\algblock[Name]{Start}{End}
		\algtext*{End}
		
		\State \textbf{Inputs:} Model $\pi_\theta$, a set of contexts and statements $G_0=\{(\context_i, \statement_i)\}_i$, maximum depth $d$.
		
		\For{$\depth \gets 0,1,\cdots,d-1$}
			\State Sample proofs $(t_{0,i}\; \cp_i)\sim \pi_\theta(\cdot\mid \context_i\;\statement_i)$ for lemmas $(\context_i, \statement_i)$ in $G_\depth$, where $t_{0,i}$ is the token representing whether the proof should use invoke and $\cp_i$ is the conditional proof.\label{line:3}
			\State $P_{\depth}\gets \{(\context_i, \statement_i, \cp_i) \mid \forall (\context_i, \statement_i)\in G_\depth\}.$
			\State Collect proposed lemmas (a conditional proof $\cp_i$ might propose more than one lemma $l_j$): 
   \begin{center}$G_{\depth+1}\gets \{(\context_i, l_j)\mid (\context_i, \statement_i, \cp_i)\in P_{\depth} \text{ and } l_j \text{ is proposed in }\cp_i\}.$\end{center}\label{line:5}
		\EndFor
		\State Sample proofs $\cp_i\sim \pi_\theta(\cdot\mid \context_i\;\statement_i\;\noinvoke)$ for $(\context_i, \statement_i)$ in $G_d$. $\Comment${ Truncate at depth $d$.} \label{line:6}
		\State $P_{d}\gets \{(\context_i, \statement_i, \cp_i) \mid \forall (\context_i, \statement_i)\in G_d\}.$
		\State \textbf{Return} $\cup_{\depth=0}^{d}P_\depth.$
	\end{algorithmic}
\end{algorithm}

\subsection{Reinforcement learning with lemma proposal}\label{sec:rl}
Our reinforcement learning method is illustrated in Fig.~\ref{fig:alg}. We start with a supervised fine-tuned model so that it can generate conditional proofs in the desired format. Then at every round, we randomly sample a batch of examples $D$ from the training dataset and perform the following steps.

\paragraph{Step 1: Generate proofs.} We first generate proof trees for every theorem in $D$.
For better exploration, we use a modified version of Alg.~\ref{alg:tree-generation-test} (shown in Alg.~\ref{alg:tree-generation-train} of Appendix~\ref{app:generation-train}) with the following differences:
\begin{enumerate}
	\item for the theorems where the probability $\pi_\theta(\useinvoke \mid x)$ is among the top 50\% in the batch, we will force the model to generate conditional proofs with $t_0 = \useinvoke$. Otherwise, we sample $t_0$ according to the probability of $\pi_\theta(\cdot \mid x)$, and
	\item for every theorem where the model generates a conditional proof with new lemmas, we also let the model generate another conditional proof without proposing lemmas. If any of these two conditional proofs is globally correct, it can be used to construct proof trees for other theorems.
\end{enumerate}

\paragraph{Step 2: Determine the reward of an example.} In this step, we first check the local correctness of each conditional proof using the formal verifier (Step 2a in Fig.~\ref{fig:alg}). 

In addition to the verifiers' output, we apply two filters to help train the model: (a) we filter out trivial lemma proposals --- if a proposed lemma directly implies the theorem (e.g., if the proposed lemma has exactly the same statement as the theorem), we simply discard this example, and (b) we remove unnecessary lemma proposals --- if the conditional proof is still correct after removing all the references to a proposed lemma, we remove this lemma from the conditional proof.

We then determine the global correctness of the generated proofs. Finally, we assign a binary reward $r(c,s,\cp)$ to each tree node with context $c$, statement $s$, and conditional proof $\cp$ based on its global correctness  (Step 2b in Fig.~\ref{fig:alg}).

\paragraph{Step 3: Update the model by REINFORCE.} In this step, we first construct a training dataset consisting of examples with format (prompt, target, weight) from the conditional proofs collected in Step 1, and then update the model $\pi_\theta$ using the weighted cross-entropy loss.

For each generated conditional proof, we add one example to the training dataset where the prompt is the context concatenated with the theorem statement, and the target is the conditional proof prepended by the \useinvoke or \noinvoke token.
Note that the reward of a conditional proof depends not only on the correctness of the conditional proof, but also on the correctness of the proposed lemmas.

To reduce the variance of our gradient updates, we train a value function $V_\phi$ that predicts the expected reward of the current policy on a given proof tree node (i.e., $V_\phi\approx \E_{\cp\sim \pi(\cdot\mid c,s)}[r(c,s,\cp)]$). The weight of an example is the product of the value function's outputs on invoked lemmas multiplied with a length penalty to incentivize shorter proofs --- for a proof tree node with conditional proof length $h$ and proposed lemmas $l_1,\cdots,l_k$, the weight $w$ of this example is
$
	w=\gamma^h\prod\nolimits_{i=1}^{k}V_\phi(l_i)
$
with discount factor $\gamma\in (0,1)$, or $w=0$ if the proof tree node is not locally correct.

To simplify the implementation, we train the value function to predict two special tokens, \true and \false, conditioned on the context and theorem statement. Let $p_{\text t}$ be the probability of the \true token conditioned on the context and theorem statement, and $p_{\text f}$ the probability of the \false token. The output of the value function is then $p_{\text t} / (p_{\text t} + p_{\text f})$.

As done for the SFT dataset, we add one augmented example by moving the proposed lemma from the conditional proof to the context for any locally correct conditional proof with new lemmas. 
We also add examples constructed from the human-written conditional proofs (i.e., the ground-truth proofs) of theorems in the batch $D$. In addition, we use a replay buffer to stabilize the training.

\paragraph{Remarks.} Note that we update the model using partial proofs (i.e., proof sub-trees) even if the original theorem from the dataset (i.e, the root of the proof tree) is not proved. Hence, our method can also be viewed as an instantiation of hindsight experience replay \citep{andrychowicz2017hindsight}, where the hindsight trajectories are correct proof sub-trees.

Our algorithm is also closely related to expert iteration. In our notation, expert iteration is equivalent to using a binary weight $w=\ind{\text{the proof tree node is \emph{globally} correct}}.$
	\section{Experiments}\label{sec:experiments}
This section presents our experimental results. We first list additional experiment details (Section~\ref{sec:exp-details}) and then compare our setup with prior works (Section~\ref{sec:setup-compare}). Finally, we show our main results in Section~\ref{sec:results} and examples of proposed lemmas in Section~\ref{sec:case-study}.
\subsection{Experiment details}\label{sec:exp-details}
\paragraph{Proof verification software.} We use Isabelle \citep{nipkow2002isabelle} as our proof verification software since the proofs are declarative and human-readable without knowing the verifier's proof state, and we use PISA (Portal to ISAbelle, \citet{jiang2021lisa}) to interact with Isabelle. To check whether a proof tree node is locally correct, we import all the theorems from its premises, move each of the proposed lemmas from the conditional proof to the context, and then add a fake proof indicated by the keyword `sorry' to every lemma statement in the context (In Isabelle, `sorry' will register the statement as a fact even without any actual proof.) The remaining proof steps will follow the original Isabelle syntax, and we can check their correctness directly. We set a 10s timeout for each proof step.

\paragraph{Datasets.} Our SFT dataset consists of theorems from Archive of Formal Proof\footnote{\url{https://www.isa-afp.org/}} (AFP, retrieved on 2022-12-06) and Isabelle built-in files (such as HOL which contains the theorems that define natural numbers, etc.). The resulting dataset contains 312k examples.

For the test datasets AFP test and AFP 2023, we only keep the theorems in $\Ttree$. To construct the test set AFP 2023, we parse the AFP files submitted after the knowledge cutoff date of our pretrained model (April 2023) to eliminate possible data leakages.\footnote{Here we use the archive of AFP retrieved on 2023-11-22.} The AFP test set contains 4.3k theorems and the AFP 2023 test set 2k theorems.

\paragraph{Testing setup.} 
To measure the performance of the models, we sample $k$ proof trees per theorem independently on the test set and report the pass@k performance (that is, a theorem is proved if at least one of the conditional proofs is globally correct with respect to all the generated tree nodes). When generating proofs, we use temperature 0.7 and truncate the context to only include the last 1k tokens. The proof trees are truncated at depth 2.

\paragraph{Supervised fine-tuning.} We start from the Llemma 7b model \citep{azerbayev2023llemma} and fine-tune the model for 2 epochs with the standard cross entropy loss.\footnote{In our preliminary experiments, we observe that the model overfits after 2 epochs}
On theorems from AFP, we compute the loss only on the special token and the proof, but not on the context and statement. On Isabelle built-in theorems, we compute the loss on the statement to help the model internalize basic facts. 
We use the AdamW optimizer \citep{loshchilov2018decoupled} with linear warmup followed by a constant learning rate 1e-5, macro batch size 128, and context window 2048.

\paragraph{Reinforcement learning.} The dataset we use for the reinforcement learning stage is $\Ttree$, the set of theorems that are iteratively peeled off when parsing AFP files --- $\Ttree$ contains 104k examples. We first train the model with 1 epoch of supervised fine-tuning, and then run RL for 20 rounds with a batch of 5k random examples per round. 
We truncate the proof tree at depth 3 and sample with temperature 0.7 during training for better exploration. We use the same hyperparameters as the SFT stage to update the policy $\pi_\theta$, 
We initialize the value function $V_\phi$ with a Llemma 7b model fine-tuned on our SFT dataset.

\subsection{Comparison of our new setup with prior works}\label{sec:setup-compare}
In this section, we concretely compare our new setup with prior works \citep{jiang2021lisa, first2023baldur}. Recall that there are two main differences in how we process our dataset:
\begin{enumerate}
\item we split the train/test set based on file dependencies so that no theorems in the test set are referred to in the training set, whereas PISA splits theorems randomly, and
\item when testing a proof, we remove certain lemmas from the context. 
\end{enumerate}

To show that our setup is indeed more challenging, we first construct datasets formatted similarly to those in \citep{first2023baldur}. Specifically, we parse the AFP files into examples using the method described in Section~\ref{sec:setup}, with the only exception being that all human-written lemmas are kept in the context. Then we select a subset of theorems as the test set $D_{\text{test}}^{\text{w/ l}}$ based on the dependencies of the AFP files so that the examples in the test set are never used by the remaining theorems (see Section~\ref{sec:hyperparameters} for more details). Then we split the remaining examples randomly into training and validation datasets, denoted by $D_{\text{train}}^{\text{w/ l}}$ and $D_{\text{val}}^{\text{w/ l}}$ respectively. We use $D_{\text{test}}^{\text{w/o l}}$ to denote the test dataset of our setup where the lemmas are removed from the context. The validation dataset $D_{\text{val}}^{\text{w/ l}}$ mimics prior works' setup \citep{jiang2021lisa,first2023baldur}, and $D_{\text{test}}^{\text{w/ l}}$ is an interpolation between prior works' setup and our setup. Table~\ref{table:setup} shows the performance of the model with supervised fine-tuning on $D_{\text{train}}^{\text{w/ l}}$ and all Isabelle built-in theorems. The results suggest that both features of our setup, removing the lemmas and splitting the training/test set by file dependencies, increase the difficulty of the task.

\begin{table}[htp]
	\centering
	\caption{Pass rate on different dataset formats and partitions of the SFT model trained on the $D_{\text{train}}^{\text{w/ l}}$. The validation dataset $D_{\text{val}}^{\text{w/ l}}$ mimics the test setup of prior works, and our setup is $D_{\text{test}}^{\text{w/o l}}$ where the same model performs much worse. The results suggest that our setup is indeed more challenging.}
	\begin{tabular}{lccc}
		\toprule
		\multirow{2}{*}{Test setup}   & $D_{\text{val}}^{\text{w/ l}}$: w/ lemmas & $D_{\text{test}}^{\text{w/ l}}$: w/ lemmas    & $D_{\text{test}}^{\text{w/o l}}$: w/o lemmas   \\
		& split randomly   & split by dependency & split by dependency \\ \midrule
		\multicolumn{1}{l}{pass@4} & 45.7             & 39.7              &  35.7               \\
		\bottomrule
	\end{tabular}
	\label{table:setup}
\end{table}

\subsection{Main results}\label{sec:results}
This section reports the models' pass@k performance on AFP test and AFP 2023 datasets. For a given dataset, pass@k measures the percentage of the theorems proved by at least one of $k$ proofs generated by the model.
Recall that the theorems in the test set are selected based on the dependencies of the AFP files and are not used in any proofs from the training set (see Section~\ref{sec:setup-compare} for more details). In other words, we do not train the model on test datasets using reinforcement learning. Instead, we test whether \method-RL is a fundamentally better model when tested on new theorems.

\begin{table}[htp]
	\centering
	\caption{Pass@16 of different models on AFP test sets. Our model with reinforcement learning (\method-RL) improves upon the SFT model and outperforms baseline methods.}
	\label{table:afp}
	\begin{tabular}{l|ccccc}
		\toprule
		Test set & \shortstack{SFT w/o \\ lemma proposal} & \shortstack{RL w/o \\ lemma proposal} & \method-SFT & \method-RL \\ \midrule
		AFP test & 43.4 & 42.4 & 40.8 & \textbf{45.5} \\ \midrule
		AFP 2023 & \textbf{39.4} & 37.7 & 36.5 & \textbf{39.5} \\
		\bottomrule
	\end{tabular}
\end{table}

As a baseline method, we train a model on a variant of the SFT dataset where all lemmas are kept in the context. It can be seen as a reproduction of \citet{first2023baldur} with a slightly different way to obtain the context --- \citet{first2023baldur} includes all the file content before the statement of the theorem, whereas we only keep the statement of previous lemmas. We also run reinforcement learning on the same RL dataset as our method (see Section~\ref{sec:baseline-details} for more details).

\begin{figure}[htp]
	\centering
	\includegraphics[width=.45\linewidth]{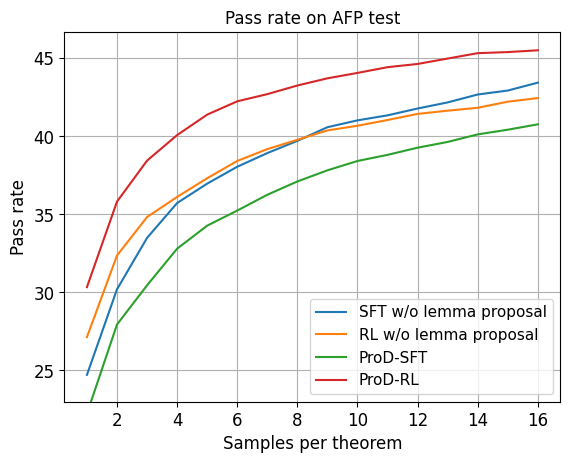}~~
	\includegraphics[width=.45\linewidth]{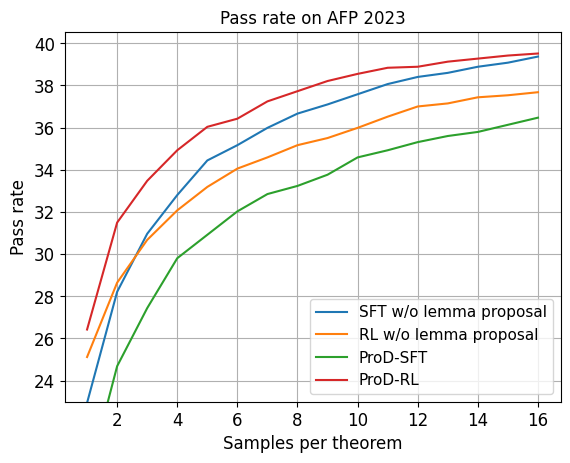}~~
	\caption{The pass rate of different models on AFP test (Left) and AFP 2023 (Right) test sets. Our RL model improves upon the SFT model whereas without proposing new lemmas (RL w/o lemma proposal), we do not observe any improvement.}
	\label{fig:main}
\end{figure}
Table~\ref{table:afp} shows the performance of our model on the AFP test sets. For a fair comparison, the baseline models are tested in our new setup without human-written lemmas. Note that the SFT model without lemma proposal outperforms the SFT model with lemma proposal. We hypothesize that it is because proposing correct lemmas itself is challenging, which distracts the model from learning to generate direct proofs. However, RL with lemma proposal improves the SFT model and outperforms others because the model proposes and proves additional lemmas that are not in the training dataset, whereas RL without lemma proposal yields no improvement. %

In Fig.~\ref{fig:main}, we plot the pass rates with different numbers of samples per theorem on both AFP test and AFP 2023. Fig.~\ref{fig:main} shows that on AFP test, the \method-RL model significantly improves upon baseline methods as well as the \method-SFT. However, on AFP 2023, the improvement is minor over SFT w/o lemma proposal, while \method-RL still outperforms \method-SFT. The results suggest that the baseline methods are more robust to heavier distribution shifts, while our method has a larger improvement when the test distribution is closer to the training distribution.
\begin{figure}[htp]
\centering
\includegraphics[width=.45\linewidth]{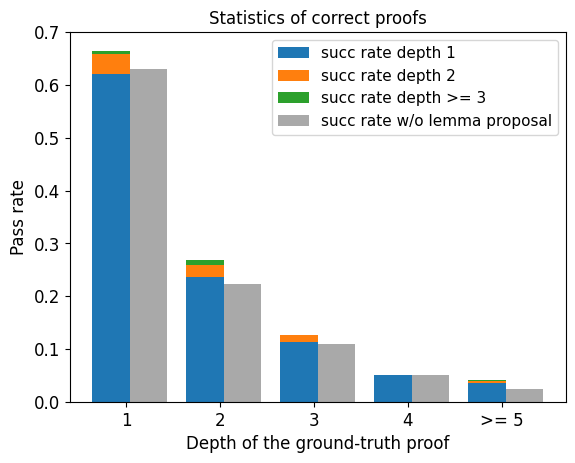}
\caption{The pass rate of theorems in AFP test grouped by the depth of their ground-truth proof. Grey bars represent the proof generated by the model SFT w/o lemma proposal, and the colored bars represent the proof trees generated by \method-RL with various depths.}
\label{fig:main-depth}
\end{figure}

In Fig.~\ref{fig:main-depth}, we decompose the proved theorems by the depth of their ground-truth proofs (shown on the $x$-axis) and the depth of generated proof trees (indicated by color). When there are multiple correct proof trees, we plot the one with the maximum depth. As a comparison, we also plot the success rates of the proofs generated by the RL model trained w/o lemma proposal. Fig.~\ref{fig:main-depth} shows that the improvement of \method-RL mostly comes from proving theorems with low-to-medium difficulty where the depth of the ground-truth proof is at most 2. For more complex theorems, both models' pass rates are low and the improvement of our method is not significant, meaning that they are currently beyond the models' capability.

\subsection{Case study of proposed lemmas}\label{sec:case-study}
In this section, we manually examine the new lemmas proposed during RL and list the typical cases where new lemmas are proposed. Note that many AFP files focus on complex concepts and results in mathematics or computer science, making manual examination challenging. Therefore, examples in this section are biased toward easier theorems.

\paragraph{Case 1: Model decomposes theorems into lemmas.} 
In this case, the model correctly decomposes the proof of a theorem into several lemmas. The following example belongs to the AFP file {\ttfamily List-Infinite}, which focuses on lists and sets with infinite size. The theorem (Line 1) states that the cardinality of the set $A\cup \{x\}$ equals $|A|$ if $x\in A$, or the successor integer of $|A|$ otherwise (i.e., $|A|+1$ for finite $A$ and $\infty$ otherwise). During the proof (Lines 2-4), our model proposes two lemmas in Lines 2 and 3 to deal with the two possible cases ($x\not\in A$ or $x\in A$) respectively. Finally, Line 4 proves the original theorem using the two proposed lemmas.
\begin{lstlisting}[language=Isabelle,basicstyle=\scriptsize\ttfamily]
theorem icard_insert_if: "icard (insert x A) = (if x \<in> A then icard A else eSuc (icard A))"
\end{lstlisting}
\vspace{-10pt}
\begin{lstlisting}[backgroundcolor=\color{lightblue},language=Isabelle,firstnumber=2,basicstyle=\scriptsize\ttfamily]
	<invoke> lemma icard_insert_disjoint: "x ∉ A ⟹ icard (insert x A) = eSuc (icard A)" </invoke>
	<invoke> lemma icard_insert_eq: "x \<in> A ⟹ icard (insert x A) = icard A" </invoke>
	by (simp add: icard_insert_eq icard_insert_disjoint)
\end{lstlisting}

\paragraph{Case 2: The proposed lemma is a rephrase of an existing lemma.} We also find that some proposed lemmas are rephrases of existing lemmas in the training dataset. Although in this case, the proposed lemma is not fundamentally useful for proving new theorems, they can be viewed as data augmentation to enhance the models' performance. In the following example, the model produces a lemma equivalent to one in an AFP file. Line 1 shows the original form of the lemma stated in the AFP file, while Lines 2-4 show an equivalent lemma proposed by our model during RL.
\begin{lstlisting}[language=Isabelle]
lemma icard_mono: "A \<subseteq> B ⟹ icard A \<le> icard B"
\end{lstlisting}
\vspace{-10pt}
\begin{lstlisting}[language=Isabelle,firstnumber=2]
lemma icard_mono:
	assumes "A \<subseteq> B"
	shows "icard A \<le> icard B"
\end{lstlisting}

\paragraph{Case 3: The proposed lemma is novel but not useful to the original proof.} We also observe cases where the proposed lemma is novel, but the conditional proof of the theorem is incorrect. In the following example, the proposed lemma states that the shortest path between vertices $u,v$ is a lower bound for the length of any path that connects $u,v$ (in an unweighted and undirected graph):
\begin{lstlisting}[language=Isabelle]
lemma shortest_path_lower_bound: 
	assumes "p \<in> connecting_paths u v"
	shows "shortest_path u v \<le> enat (walk_length p)"
\end{lstlisting}
This lemma is proposed to prove that the shortest path between the vertex $u$ and itself has length 0 (which is a theorem in the AFP file). However, the conditional proof of the theorem contains a few mistakes while the proposed lemma is proved separately. In this case, we still train on the correct lemma even though it might not be directly useful to the theorem in the training set.

\paragraph{Remarks.}%
We observe that the lemmas proposed by the model typically do not involve complex ideas. We attribute this to two main factors: (a) the limited size of our model and formal proof dataset, and (b) the fact that many human-written lemmas in the AFP file are indeed about basic facts and basic properties (which are often used to prove more complex theorems later). Nevertheless, our model still proposes and proves reasonable lemmas that are not present in the training dataset, and our experiments demonstrate that with these proposed lemmas, \method-RL outperforms \method-SFT on holdout test sets. We leave it to future work to scale up our method and force the model to focus on more challenging theorems.
	\section{Related works}

To generate formal proofs with language models, most prior methods provide the verifier’s state to the model to sample the proofs step-by-step, and use algorithms like MCTS to search for a correct complete proof \citep{polu2020generative,jiang2021lisa,han2021proof,polu2022formal,jiang2022thor,lample2022hypertree,wang2024proving,yang2024leandojo}. The major drawback of these methods is their high computation cost at test time.
Several recent works \citep{first2023baldur,xin2024deepseeka, xin2024deepseekb} use large language models to generate a whole proof directly without the verifier’s state. Our baseline, the SFT model without lemma proposal, can be viewed as a reproduction of \citet{first2023baldur} with a slightly different way of computing the context.

Prior works also use reinforcement learning or expert iteration to improve the models’ performance on writing formal proofs, where the training datasets contain formal synthetic inequalities \citep{polu2022formal} or statements translated from natural language mathematical problems \citep{wu2022autoformalization, xin2024deepseeka, xin2024deepseekb}. In contrast, we aim to improve the models' performance without any additional (even unlabeled) data. As future works, one can extend our algorithms to more diverse additional datasets.

Another line of research aims to translate natural language proofs into formal proofs \citep{jiang2022draft,zheng2023lyra}. \citet{xin2023lego} build a library of useful lemmas by decomposing natural language proofs into lemmas with an LLM and then formalizing the decomposed proofs. In contrast, we propose new lemmas entirely in formal language.

Automated theorem provers (ATPs) have been extensively studied, with various learning- or search-based methods developed to generate tactics for a given proof state (e.g., \citet{gauthier2021tactictoe,SCV:CADE-2019} and references therein). In comparison, our method focuses on generating multi-step proofs with LLMs while using existing ATP tools to complete low-level details. Orthogonally, a recent method \citep{mikula2023magnushammer} improves existing provers with a transformer-based retrieval model as the premise-selection tool, and could potentially be combined with our methods.

In general, mathematical question-answering tasks (such as GSM8K \citep{cobbe2021training} and MATH \citep{hendrycks2021measuring}) and theorem-proving tasks (such as \citet{welleck2021naturalproofs}) are well-accepted benchmarks for the reasoning capability of large language models. Prior works show that instruction tuning or RL can significantly improve the models’ performance \citep{shao2024deepseekmath}. However, evaluation on these tasks is either performed by another language model (which is prone to errors) \citep{lightman2023let}, or requires ground-truth answers that are hard to acquire at scale.
	\section{Conclusion}

In this paper, we design a reinforcement learning algorithm that encourages LLMs to write formal proofs by decomposing them hierarchically. We also design a more natural testing setup by removing the directly relevant lemmas from the context. We show that, by proposing and proving new lemmas that are not present in the training dataset, the resulting model \method-RL outperforms or achieves comparable performance to baseline methods trained on the same dataset.

\paragraph{Limitation.} We observe that the improvement of \method-RL over SFT w/o lemma proposal is significant only when the test distribution is close to the training distribution. On miniF2F \citep{zheng2021minif2f} where the theorems are very different from theorems in the training dataset, \method-RL performs worse than SFT w/o lemma proposal, as shown in Table~\ref{table:miniF2F}. We also observe that when tested on miniF2F theorems, our model failed to propose meaningful lemmas. This may be because proving to miniF2F-level mathematics questions typically does not require hierarchical decomposition. Therefore we leave it as future work to extend our methods to other domains such as miniF2F and PutnamBench \citep{tsoukalas2024putnambench}.
\begin{table}[htp]
	\centering
	\caption{Pass@64 of different models on miniF2F. \method-RL performs worse than SFT w/o lemma proposal.}
	\label{table:miniF2F}
	\begin{tabular}{l|ccccc}
		\toprule
		Test set & \shortstack{SFT w/o \\ lemma proposal} & \shortstack{RL w/o \\ lemma proposal} & \method-SFT & \method-RL \\ \midrule
		miniF2F valid & \textbf{46.3} & 40.6 & 44.7 & {41.4} \\ \midrule
		miniF2F test & \textbf{40.6} & 38.9 & 39.3 & {39.3} \\
		\bottomrule
	\end{tabular}
\end{table}
	\subsection*{Acknowledgment}
	The authors would like to thank Neil Band, Zhizhou Ren, and Yuanhao Wang for their helpful discussions. The authors would also like to thank the support from NSF CIF 2212263.
	
\bibliographystyle{plainnat}
\bibliography{all.bib, new.bib}

\begin{thebibliography}{41}
\providecommand{\natexlab}[1]{#1}
\providecommand{\url}[1]{\texttt{#1}}
\expandafter\ifx\csname urlstyle\endcsname\relax
  \providecommand{\doi}[1]{doi: #1}\else
  \providecommand{\doi}{doi: \begingroup \urlstyle{rm}\Url}\fi

\bibitem[Andrychowicz et~al.(2017)Andrychowicz, Wolski, Ray, Schneider, Fong,
  Welinder, McGrew, Tobin, Abbeel, and Zaremba]{andrychowicz2017hindsight}
Marcin Andrychowicz, Filip Wolski, Alex Ray, Jonas Schneider, Rachel Fong,
  Peter Welinder, Bob McGrew, Josh Tobin, Pieter Abbeel, and Wojciech Zaremba.
\newblock Hindsight experience replay.
\newblock In \emph{Proceedings of the 31st International Conference on Neural
  Information Processing Systems}, pages 5055--5065, 2017.

\bibitem[Azerbayev et~al.(2023)Azerbayev, Schoelkopf, Paster, Dos~Santos,
  McAleer, Jiang, Deng, Biderman, and Welleck]{azerbayev2023llemma}
Zhangir Azerbayev, Hailey Schoelkopf, Keiran Paster, Marco Dos~Santos, Stephen
  McAleer, Albert Jiang, Jia Deng, Stella Biderman, and Sean Welleck.
\newblock Llemma: An open language model for mathematics.
\newblock In \emph{The 3rd Workshop on Mathematical Reasoning and AI at
  NeurIPS'23}, 2023.

\bibitem[Burns et~al.(2023)Burns, Izmailov, Kirchner, Baker, Gao,
  Aschenbrenner, Chen, Ecoffet, Joglekar, Leike, et~al.]{burns2023weak}
Collin Burns, Pavel Izmailov, Jan~Hendrik Kirchner, Bowen Baker, Leo Gao,
  Leopold Aschenbrenner, Yining Chen, Adrien Ecoffet, Manas Joglekar, Jan
  Leike, et~al.
\newblock Weak-to-strong generalization: Eliciting strong capabilities with
  weak supervision.
\newblock \emph{arXiv preprint arXiv:2312.09390}, 2023.

\bibitem[Church(1936)]{church1936note}
Alonzo Church.
\newblock A note on the entscheidungsproblem.
\newblock \emph{The journal of symbolic logic}, 1\penalty0 (1):\penalty0
  40--41, 1936.

\bibitem[Cobbe et~al.(2021)Cobbe, Kosaraju, Bavarian, Chen, Jun, Kaiser,
  Plappert, Tworek, Hilton, Nakano, et~al.]{cobbe2021training}
Karl Cobbe, Vineet Kosaraju, Mohammad Bavarian, Mark Chen, Heewoo Jun, Lukasz
  Kaiser, Matthias Plappert, Jerry Tworek, Jacob Hilton, Reiichiro Nakano,
  et~al.
\newblock Training verifiers to solve math word problems.
\newblock \emph{arXiv preprint arXiv:2110.14168}, 2021.

\bibitem[De~Moura et~al.(2015)De~Moura, Kong, Avigad, Van~Doorn, and von
  Raumer]{de2015lean}
Leonardo De~Moura, Soonho Kong, Jeremy Avigad, Floris Van~Doorn, and Jakob von
  Raumer.
\newblock The lean theorem prover (system description).
\newblock In \emph{Automated Deduction-CADE-25: 25th International Conference
  on Automated Deduction, Berlin, Germany, August 1-7, 2015, Proceedings 25},
  pages 378--388. Springer, 2015.

\bibitem[First et~al.(2023)First, Rabe, Ringer, and Brun]{first2023baldur}
Emily First, Markus Rabe, Talia Ringer, and Yuriy Brun.
\newblock Baldur: Whole-proof generation and repair with large language models.
\newblock In \emph{Proceedings of the 31st ACM Joint European Software
  Engineering Conference and Symposium on the Foundations of Software
  Engineering}, pages 1229--1241, 2023.

\bibitem[Fleming et~al.(2023)Fleming, Lozano, Haberkorn, Jindal, Reis, Thapa,
  Blankemeier, Genkins, Steinberg, Nayak, et~al.]{fleming2023medalign}
Scott~L Fleming, Alejandro Lozano, William~J Haberkorn, Jenelle~A Jindal,
  Eduardo~P Reis, Rahul Thapa, Louis Blankemeier, Julian~Z Genkins, Ethan
  Steinberg, Ashwin Nayak, et~al.
\newblock Medalign: A clinician-generated dataset for instruction following
  with electronic medical records.
\newblock \emph{arXiv preprint arXiv:2308.14089}, 2023.

\bibitem[Gauthier et~al.(2021)Gauthier, Kaliszyk, Urban, Kumar, and
  Norrish]{gauthier2021tactictoe}
Thibault Gauthier, Cezary Kaliszyk, Josef Urban, Ramana Kumar, and Michael
  Norrish.
\newblock Tactictoe: learning to prove with tactics.
\newblock \emph{Journal of Automated Reasoning}, 65\penalty0 (2):\penalty0
  257--286, 2021.

\bibitem[Guha et~al.(2024)Guha, Nyarko, Ho, R{\'e}, Chilton, Chohlas-Wood,
  Peters, Waldon, Rockmore, Zambrano, et~al.]{guha2024legalbench}
Neel Guha, Julian Nyarko, Daniel Ho, Christopher R{\'e}, Adam Chilton, Alex
  Chohlas-Wood, Austin Peters, Brandon Waldon, Daniel Rockmore, Diego Zambrano,
  et~al.
\newblock Legalbench: A collaboratively built benchmark for measuring legal
  reasoning in large language models.
\newblock \emph{Advances in Neural Information Processing Systems}, 36, 2024.

\bibitem[Han et~al.(2021)Han, Rute, Wu, Ayers, and Polu]{han2021proof}
Jesse~Michael Han, Jason Rute, Yuhuai Wu, Edward Ayers, and Stanislas Polu.
\newblock Proof artifact co-training for theorem proving with language models.
\newblock In \emph{International Conference on Learning Representations}, 2021.

\bibitem[Hendrycks et~al.(2021)Hendrycks, Burns, Kadavath, Arora, Basart, Tang,
  Song, and Steinhardt]{hendrycks2021measuring}
Dan Hendrycks, Collin Burns, Saurav Kadavath, Akul Arora, Steven Basart, Eric
  Tang, Dawn Song, and Jacob Steinhardt.
\newblock Measuring mathematical problem solving with the math dataset.
\newblock In \emph{Thirty-fifth Conference on Neural Information Processing
  Systems Datasets and Benchmarks Track (Round 2)}, 2021.

\bibitem[Jiang et~al.(2022{\natexlab{a}})Jiang, Welleck, Zhou, Li, Liu, Jamnik,
  Lacroix, Wu, and Lample]{jiang2022draft}
Albert~Q Jiang, Sean Welleck, Jin~Peng Zhou, Wenda Li, Jiacheng Liu, Mateja
  Jamnik, Timoth{\'e}e Lacroix, Yuhuai Wu, and Guillaume Lample.
\newblock Draft, sketch, and prove: Guiding formal theorem provers with
  informal proofs.
\newblock \emph{arXiv preprint arXiv:2210.12283}, 2022{\natexlab{a}}.

\bibitem[Jiang et~al.(2021)Jiang, Li, Han, and Wu]{jiang2021lisa}
Albert~Qiaochu Jiang, Wenda Li, Jesse~Michael Han, and Yuhuai Wu.
\newblock Lisa: Language models of isabelle proofs.
\newblock In \emph{6th Conference on Artificial Intelligence and Theorem
  Proving}, pages 378--392, 2021.

\bibitem[Jiang et~al.(2022{\natexlab{b}})Jiang, Li, Tworkowski, Czechowski,
  Odrzyg{\'o}{\'z}d{\'z}, Mi{\l}o{\'s}, Wu, and Jamnik]{jiang2022thor}
Albert~Qiaochu Jiang, Wenda Li, Szymon Tworkowski, Konrad Czechowski, Tomasz
  Odrzyg{\'o}{\'z}d{\'z}, Piotr Mi{\l}o{\'s}, Yuhuai Wu, and Mateja Jamnik.
\newblock Thor: Wielding hammers to integrate language models and automated
  theorem provers.
\newblock \emph{Advances in Neural Information Processing Systems},
  35:\penalty0 8360--8373, 2022{\natexlab{b}}.

\bibitem[Lample et~al.(2022)Lample, Lacroix, Lachaux, Rodriguez, Hayat, Lavril,
  Ebner, and Martinet]{lample2022hypertree}
Guillaume Lample, Timothee Lacroix, Marie-Anne Lachaux, Aurelien Rodriguez,
  Amaury Hayat, Thibaut Lavril, Gabriel Ebner, and Xavier Martinet.
\newblock Hypertree proof search for neural theorem proving.
\newblock \emph{Advances in neural information processing systems},
  35:\penalty0 26337--26349, 2022.

\bibitem[Lightman et~al.(2023)Lightman, Kosaraju, Burda, Edwards, Baker, Lee,
  Leike, Schulman, Sutskever, and Cobbe]{lightman2023let}
Hunter Lightman, Vineet Kosaraju, Yura Burda, Harri Edwards, Bowen Baker, Teddy
  Lee, Jan Leike, John Schulman, Ilya Sutskever, and Karl Cobbe.
\newblock Let's verify step by step.
\newblock \emph{arXiv preprint arXiv:2305.20050}, 2023.

\bibitem[Loshchilov and Hutter(2018)]{loshchilov2018decoupled}
Ilya Loshchilov and Frank Hutter.
\newblock Decoupled weight decay regularization.
\newblock In \emph{International Conference on Learning Representations}, 2018.

\bibitem[Miku{\l}a et~al.(2023)Miku{\l}a, Antoniak, Tworkowski, Jiang, Zhou,
  Szegedy, Kuci{\'n}ski, Mi{\l}o{\'s}, and Wu]{mikula2023magnushammer}
Maciej Miku{\l}a, Szymon Antoniak, Szymon Tworkowski, Albert~Qiaochu Jiang,
  Jin~Peng Zhou, Christian Szegedy, {\L}ukasz Kuci{\'n}ski, Piotr Mi{\l}o{\'s},
  and Yuhuai Wu.
\newblock Magnushammer: A transformer-based approach to premise selection.
\newblock \emph{arXiv preprint arXiv:2303.04488}, 2023.

\bibitem[M{\"u}ndler et~al.(2023)M{\"u}ndler, He, Jenko, and
  Vechev]{mundler2023self}
Niels M{\"u}ndler, Jingxuan He, Slobodan Jenko, and Martin Vechev.
\newblock Self-contradictory hallucinations of large language models:
  Evaluation, detection and mitigation.
\newblock In \emph{The Twelfth International Conference on Learning
  Representations}, 2023.

\bibitem[Nipkow et~al.(2002)Nipkow, Wenzel, and Paulson]{nipkow2002isabelle}
Tobias Nipkow, Markus Wenzel, and Lawrence~C Paulson.
\newblock \emph{Isabelle/HOL: a proof assistant for higher-order logic}.
\newblock Springer, 2002.

\bibitem[Paulsson and Blanchette(2012)]{paulsson2012three}
Lawrence~C Paulsson and Jasmin~C Blanchette.
\newblock Three years of experience with sledgehammer, a practical link between
  automatic and interactive theorem provers.
\newblock In \emph{Proceedings of the 8th International Workshop on the
  Implementation of Logics (IWIL-2010), Yogyakarta, Indonesia. EPiC}, volume~2,
  2012.

\bibitem[Polu and Sutskever(2020)]{polu2020generative}
Stanislas Polu and Ilya Sutskever.
\newblock Generative language modeling for automated theorem proving.
\newblock \emph{arXiv preprint arXiv:2009.03393}, 2020.

\bibitem[Polu et~al.(2022)Polu, Han, Zheng, Baksys, Babuschkin, and
  Sutskever]{polu2022formal}
Stanislas Polu, Jesse~Michael Han, Kunhao Zheng, Mantas Baksys, Igor
  Babuschkin, and Ilya Sutskever.
\newblock Formal mathematics statement curriculum learning.
\newblock \emph{arXiv preprint arXiv:2202.01344}, 2022.

\bibitem[Ruan et~al.(2024)Ruan, Nie, Steenbergen, He, Zhang, Guo, Liu,
  Dang~Nguyen, Wang, Ying, et~al.]{ruan2024reinforcement}
Sherry Ruan, Allen Nie, William Steenbergen, Jiayu He, JQ~Zhang, Meng Guo, Yao
  Liu, Kyle Dang~Nguyen, Catherine~Y Wang, Rui Ying, et~al.
\newblock Reinforcement learning tutor better supported lower performers in a
  math task.
\newblock \emph{Machine Learning}, pages 1--26, 2024.

\bibitem[Schulz et~al.(2019)Schulz, Cruanes, and Vukmirovi{\'c}]{SCV:CADE-2019}
Stephan Schulz, Simon Cruanes, and Petar Vukmirovi{\'c}.
\newblock Faster, higher, stronger: {E} 2.3.
\newblock In Pascal Fontaine, editor, \emph{Proc.\ of the 27th CADE, Natal,
  Brasil}, number 11716 in LNAI, pages 495--507. Springer, 2019.

\bibitem[Shao et~al.(2024)Shao, Wang, Zhu, Xu, Song, Zhang, Li, Wu, and
  Guo]{shao2024deepseekmath}
Zhihong Shao, Peiyi Wang, Qihao Zhu, Runxin Xu, Junxiao Song, Mingchuan Zhang,
  YK~Li, Y~Wu, and Daya Guo.
\newblock Deepseekmath: Pushing the limits of mathematical reasoning in open
  language models.
\newblock \emph{arXiv preprint arXiv:2402.03300}, 2024.

\bibitem[Singhal et~al.(2023)Singhal, Azizi, Tu, Mahdavi, Wei, Chung, Scales,
  Tanwani, Cole-Lewis, Pfohl, et~al.]{singhal2023large}
Karan Singhal, Shekoofeh Azizi, Tao Tu, S~Sara Mahdavi, Jason Wei, Hyung~Won
  Chung, Nathan Scales, Ajay Tanwani, Heather Cole-Lewis, Stephen Pfohl, et~al.
\newblock Large language models encode clinical knowledge.
\newblock \emph{Nature}, 620\penalty0 (7972):\penalty0 172--180, 2023.

\bibitem[Tsoukalas et~al.(2024)Tsoukalas, Lee, Jennings, Xin, Ding, Jennings,
  Thakur, and Chaudhuri]{tsoukalas2024putnambench}
George Tsoukalas, Jasper Lee, John Jennings, Jimmy Xin, Michelle Ding, Michael
  Jennings, Amitayush Thakur, and Swarat Chaudhuri.
\newblock Putnambench: Evaluating neural theorem-provers on the putnam
  mathematical competition.
\newblock \emph{arXiv preprint arXiv:2407.11214}, 2024.

\bibitem[Turing et~al.(1936)]{turing1936computable}
Alan~Mathison Turing et~al.
\newblock On computable numbers, with an application to the
  entscheidungsproblem.
\newblock \emph{J. of Math}, 58\penalty0 (345-363):\penalty0 5, 1936.

\bibitem[Valmeekam et~al.(2023)Valmeekam, Marquez, Sreedharan, and
  Kambhampati]{valmeekam2023planning}
Karthik Valmeekam, Matthew Marquez, Sarath Sreedharan, and Subbarao
  Kambhampati.
\newblock On the planning abilities of large language models-a critical
  investigation.
\newblock \emph{Advances in Neural Information Processing Systems},
  36:\penalty0 75993--76005, 2023.

\bibitem[Wang et~al.(2024)Wang, Xin, Liu, Li, Huang, Lu, Yang, Tang, Yin, Li,
  et~al.]{wang2024proving}
Haiming Wang, Huajian Xin, Zhengying Liu, Wenda Li, Yinya Huang, Jianqiao Lu,
  Zhicheng Yang, Jing Tang, Jian Yin, Zhenguo Li, et~al.
\newblock Proving theorems recursively.
\newblock \emph{arXiv preprint arXiv:2405.14414}, 2024.

\bibitem[Wang and Demszky(2023)]{wang2023chatgpt}
Rose Wang and Dorottya Demszky.
\newblock Is chatgpt a good teacher coach? measuring zero-shot performance for
  scoring and providing actionable insights on classroom instruction.
\newblock In \emph{The 61st Annual Meeting Of The Association For Computational
  Linguistics}, 2023.

\bibitem[Welleck et~al.(2021)Welleck, Liu, Le~Bras, Hajishirzi, Choi, and
  Cho]{welleck2021naturalproofs}
Sean Welleck, Jiacheng Liu, Ronan Le~Bras, Hannaneh Hajishirzi, Yejin Choi, and
  Kyunghyun Cho.
\newblock Naturalproofs: Mathematical theorem proving in natural language.
\newblock In \emph{Thirty-fifth Conference on Neural Information Processing
  Systems Datasets and Benchmarks Track (Round 1)}, 2021.

\bibitem[Wu et~al.(2022)Wu, Jiang, Li, Rabe, Staats, Jamnik, and
  Szegedy]{wu2022autoformalization}
Yuhuai Wu, Albert~Qiaochu Jiang, Wenda Li, Markus Rabe, Charles Staats, Mateja
  Jamnik, and Christian Szegedy.
\newblock Autoformalization with large language models.
\newblock \emph{Advances in Neural Information Processing Systems},
  35:\penalty0 32353--32368, 2022.

\bibitem[Xin et~al.(2023)Xin, Wang, Zheng, Li, Liu, Cao, Huang, Xiong, Shi,
  Xie, et~al.]{xin2023lego}
Huajian Xin, Haiming Wang, Chuanyang Zheng, Lin Li, Zhengying Liu, Qingxing
  Cao, Yinya Huang, Jing Xiong, Han Shi, Enze Xie, et~al.
\newblock Lego-prover: Neural theorem proving with growing libraries.
\newblock \emph{arXiv preprint arXiv:2310.00656}, 2023.

\bibitem[Xin et~al.(2024{\natexlab{a}})Xin, Guo, Shao, Ren, Zhu, Liu, Ruan, Li,
  and Liang]{xin2024deepseeka}
Huajian Xin, Daya Guo, Zhihong Shao, Zhizhou Ren, Qihao Zhu, Bo~Liu, Chong
  Ruan, Wenda Li, and Xiaodan Liang.
\newblock Deepseek-prover: Advancing theorem proving in llms through
  large-scale synthetic data.
\newblock \emph{arXiv preprint arXiv:2405.14333}, 2024{\natexlab{a}}.

\bibitem[Xin et~al.(2024{\natexlab{b}})Xin, Ren, Song, Shao, Zhao, Wang, Liu,
  Zhang, Lu, Du, et~al.]{xin2024deepseekb}
Huajian Xin, ZZ~Ren, Junxiao Song, Zhihong Shao, Wanjia Zhao, Haocheng Wang,
  Bo~Liu, Liyue Zhang, Xuan Lu, Qiushi Du, et~al.
\newblock Deepseek-prover-v1. 5: Harnessing proof assistant feedback for
  reinforcement learning and monte-carlo tree search.
\newblock \emph{arXiv preprint arXiv:2408.08152}, 2024{\natexlab{b}}.

\bibitem[Yang et~al.(2024)Yang, Swope, Gu, Chalamala, Song, Yu, Godil, Prenger,
  and Anandkumar]{yang2024leandojo}
Kaiyu Yang, Aidan Swope, Alex Gu, Rahul Chalamala, Peiyang Song, Shixing Yu,
  Saad Godil, Ryan~J Prenger, and Animashree Anandkumar.
\newblock Leandojo: Theorem proving with retrieval-augmented language models.
\newblock \emph{Advances in Neural Information Processing Systems}, 36, 2024.

\bibitem[Zheng et~al.(2023)Zheng, Wang, Xie, Liu, Sun, Xin, Shen, Li, and
  Li]{zheng2023lyra}
Chuanyang Zheng, Haiming Wang, Enze Xie, Zhengying Liu, Jiankai Sun, Huajian
  Xin, Jianhao Shen, Zhenguo Li, and Yu~Li.
\newblock Lyra: Orchestrating dual correction in automated theorem proving.
\newblock \emph{arXiv preprint arXiv:2309.15806}, 2023.

\bibitem[Zheng et~al.(2021)Zheng, Han, and Polu]{zheng2021minif2f}
Kunhao Zheng, Jesse~Michael Han, and Stanislas Polu.
\newblock minif2f: a cross-system benchmark for formal olympiad-level
  mathematics.
\newblock In \emph{International Conference on Learning Representations}, 2021.

\end{thebibliography}

\newpage
\appendix
\section{Additional experiments details}

\subsection{Generating proof trees for RL}\label{app:generation-train}
In Alg.~\ref{alg:tree-generation-train}, we present the algorithm for generating proof trees during RL training. Recall that, compared with Alg.~\ref{alg:tree-generation-test}, there are two major differences:
\begin{enumerate}
	\item for the theorems where the probability $\pi_\theta(\useinvoke \mid x)$ is among the top 50\% in the batch, we will force the model to generate conditional proofs with $t_0 = \useinvoke$ (Line~\ref{line:train-4}-\ref{line:train-6}), and
	\item for every theorem where the model generates a conditional proof with new lemmas, we also let the model generate another conditional proof without proposing new lemmas (Line~\ref{line:train-11}).
\end{enumerate}

\begin{algorithm}[t]
	\caption{Generate proof trees (train)}
	\label{alg:tree-generation-train}
	\begin{algorithmic}[1]
		\algblock[Name]{Start}{End}
		\algtext*{End}
		
		\State \textbf{Inputs:} Model $\pi_\theta$, theorems (represented by tuples of context, statement, and condition proof) $G=\{(\context_i, \statement_i, \cp_i^\star)\}_i$, maximum depth $d$.
		
		\For{$\depth \gets 0,1,\cdots,d$}
		\State Compute the invoke probability $\forall i,\; p_i=\frac{\pi_\theta(\useinvoke \mid \context_i, \statement_i)}{\pi_\theta(\useinvoke \mid \context_i, \statement_i)+\pi_\theta(\noinvoke \mid \context_i, \statement_i)}.$
		\State Let $\kappa$ be the 50\% quantile of $\{p_i\}_i$. \label{line:train-4}
		\If{$\depth < d$}
		\State $\Ginvoke=\{(\context_i, \statement_i, \cp_i^\star)\mid p_i\ge \kappa \text{ or } u_i< p_i \text{ where }u_i\sim \unif[0,1]\}.$ \label{line:train-6}
		\Else
		\State $\Ginvoke=\emptyset$.
		\EndIf
		\State Sample proofs $\cp_i\sim \pi_\theta(\cdot\mid \context_i\;\statement_i\;\noinvoke)$ for lemmas $(\context_i, \statement_i, \cp_i^\star)$ in $G$.
		\State Sample proofs $\widehat{\cp}_i\sim \pi_\theta(\cdot\mid \context_i\;\statement_i\;\useinvoke)$ for lemmas $(\context_i, \statement_i, \cp_i^\star)$ in $\Ginvoke$.
		\State $P_{\depth}\gets \{(\context_i, \statement_i, \cp_i) \mid \forall i \text{ s.t. } (\context_i, \statement_i, \cp_i^\star)\in G\} \cup \{(\context_i, \statement_i, \widehat{\cp}_i)\mid \forall i \text{ s.t. } (\context_i, \statement_i, \cp_i^\star)\in \Ginvoke\}.$ \label{line:train-11}
		\If{$\depth = 0$}
		\State $P_{\depth}\gets P_{\depth}\cup G.$ $\Comment${ In training, we also complete proof trees for ground-truth proofs.}
		\EndIf
		\State Extract proposed lemmas (note that a condition proof $\cp$ might propose more than one lemma $l$): $G\gets \{(\context, l, \Null)\mid (\context, \statement, \bar{\cp})\in P_{\depth} \text{ and } l \text{ is proposed in }\bar{\cp}\}.$
		\EndFor
		\State \textbf{Return} $\cup_{\depth=0}^{d}P_\depth.$
	\end{algorithmic}
\end{algorithm}

\subsection{Training details of baseline models} \label{sec:baseline-details}
In this section, we describe the additional details for training the baseline models using reinforcement learning.

Our RL training pipeline for the baseline models is similar to that of \method-RL, except that the models only generate proofs without lemma proposal. For RL baselines, we use the same dataset and the same hyperparameters as our method. To mix the ground-truth conditional proofs with generated proofs, we convert the conditional proofs to proofs without lemma proposal by moving all the proposed lemma in the conditional proof to the context. 

\subsection{Additional experiment details} \label{sec:hyperparameters}

\paragraph{Details of using sledgehammer in the proof.} Sledgehammer is a premise selection tool that can automatically generate proofs to solve the current goal. Although sledgehammer is not always applicable, \citet{jiang2022thor} shows that letting the model to call sledgehammer whenever it is applicable greatly improves the model's performance.

To let the model use sledgehammer, we replace the actual proof steps in the training dataset by a call to sledgehammer if the proof step either (a) contains the proof tactics `meson, metis, and smt' (these tactics are typically generated by sledgehammer), or (b) belongs to a predefined simple set of proof tactics that can be easily generated. In particular, they are 
\begin{verbatim}
	[by auto, by simp, by blast, by fastforce, by force, 
	by eval, by presburger, by sos, by arith, by linarith, 
	by (auto simp: field_simps)]
\end{verbatim}

When testing a generated proof with calls to sledgehammer, we follow the pipeline of \citep{jiang2022draft} --- first, we try to replace the `sledgehammer' command by one of the predefined tactics. If all the attempts fail, we call the actual premises selection tool in Isabelle with a 10s timeout. If the tool does not return a valid proof, we consider this step incorrect.

Note that \citet{jiang2022thor} decide when to replace the actual proof step by a call to sledgehammer more aggressively. They attempt to call sledgehammer at every proof step, and replace the actual proof step with sledgehammer if the attempt is successful. In contrast, our decision is made without interacting with the formal verifier. This is because applying sledgehammer to every proof step requires a lot of compute, which would significantly slow down the reinforcement learning process.

\paragraph{Dataset split.} Here we describe how to split the training and test data based on the dependency of the AFP files. We first compute the dependency graph by crawling the AFP website \url{https://www.isa-afp.org/entries/}, which lists the dependency of the AFP entries. Then we find the set of AFP entries that all other entries do not depend on using the dependency graph, in which we randomly sample 10\% of the entries as the holdout test set. The resulting holdout entries are:
\begin{verbatim}
[Verified_SAT_Based_AI_Planning, SIFPL, Khovanskii_Theorem, 
Bondy, Rewriting_Z, Decreasing-Diagrams-II, Registers, 
LocalLexing, FeatherweightJava, FFT, Knot_Theory, Eval_FO, 
Saturation_Framework_Extensions, Hales_Jewett, SPARCv8, 
CoSMeDis, LP_Duality, PAPP_Impossibility, Groebner_Macaulay, 
Abstract-Hoare-Logics, PCF, Jordan_Hoelder, Knights_Tour, 
FOL_Seq_Calc3, Cartan_FP, InformationFlowSlicing_Inter, LOFT, 
Diophantine_Eqns_Lin_Hom, Dynamic_Tables, Schutz_Spacetime, 
Elliptic_Curves_Group_Law, ArrowImpossibilityGS, 
Goodstein_Lambda, XML, GenClock, Topological_Semantics].
\end{verbatim}

\paragraph{Additional training detail.} We use the Llemma code base (\url{https://github.com/EleutherAI/math-lm}) for finetuning and updating the model in reinforcement learning. The discount factor used to compute the weight is $\gamma=\exp(-0.0005).$

\subsection{Compute resources} \label{sec:resources}
For supervised finetuning and reinforcement learning, we use a machine with 8 A100-80G GPUs. It takes approximately 8 GPU days in total (i.e., 1 day wall-clock time on a single machine with 8 GPUs) to finetune a 7B model on 300k examples for 2 epochs. It takes approximately 30 GPU hours to run a single RL experiment.

To generate proofs using the trained model, we use a mix of A100-80G and A5000 GPUs. On 8 A5000 GPUs, generating proof trees of depth 2 for 4k test examples takes about 1-2 hours, depending on the length of the proof and the number of proposed lemmas.

\subsection{Licenses for existing assets} \label{sec:licenses}
In this section, we list the licenses for existing assets used in this paper.
\begin{itemize}
	\item LLemma \citep{azerbayev2023llemma}: Llama 2 Community License Agreement
	\item Archive of Formal Proofs: GNU LGPL
	\item Portal to ISAbelle \citep{jiang2021lisa}: BSD 3-Clause License
	\item Isabelle \citep{nipkow2002isabelle}: BSD licenses
	\item miniF2F \citep{zheng2021minif2f}: MIT License
\end{itemize}
\section{Additional results}
In this section, we present additional experimental results.
\subsection{The effect of sampling temperatures} \label{sec:temeprature}
\begin{figure}[htp]
	\centering
	\includegraphics[width=.5\linewidth]{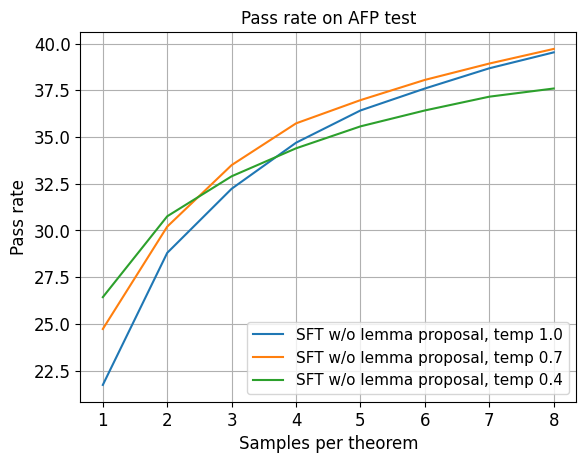}
	\caption{Pass rate of the SFT model without lemma proposal tested with different sampling temperature. We observe that lower temperature leads to better performance with 1 sample per theorem, and mildly larger temperature have better performance with more samples.}
	\label{fig:temperature}
\end{figure}

In our preliminary experiments, We tune the sampling temperature using the models trained on the AFP training sets $D_{\text{train}}^{\text{w/ l}}$ and $D_{\text{train}}^{\text{w/o l}}$ (that is, training sets constructed with and without helper lemmas, respectively). We test the model on the AFP test set $D_{\text{test}}^{\text{w/o l}}$ with different temperatures to decide the best choice for testing our models. Fig.~\ref{fig:temperature} shows the performance of the SFT model without lemma proposal using different sampling temperatures. We conclude that the temperature 0.7 is best for testing both models.

\subsection{Varying the size of the RL dataset} \label{sec:training-data-size}
\begin{figure}[htp]
	\centering
	\includegraphics[width=.5\linewidth]{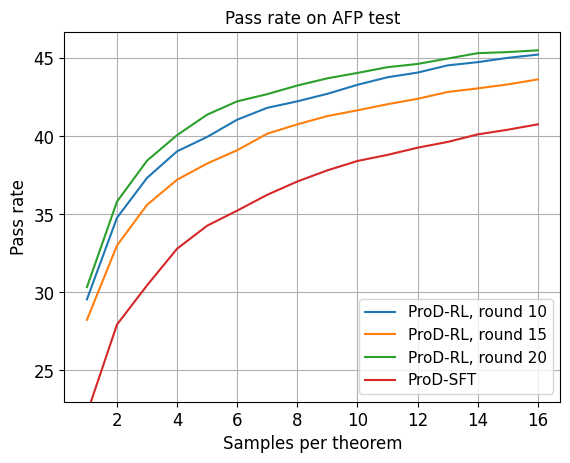}
	\caption{Pass rate of models trained with different rounds of RL training. Recall that in each round we generate proofs to a batch of 5k examples. The model trained with 20 rounds of RL achieves the best performance, and all RL models outperforms the baseline \method-SFT model.}
	\label{fig:RL-dataset-size}
\end{figure}

In this section we report the performance of the \method-RL models trained with different rounds. Recall that in each round we generate proofs to a batch of 5k examples. Therefore equivalently, Figure~\ref{fig:RL-dataset-size} shows the performance of \method-RL with a smaller RL dataset.
Note that although the performance is not monotone with respect to the RL dataset size (e.g., \method-RL at round 15 is worse than \method-RL at round 10), which might due to training instability, all the RL models significantly outperforms the baseline \method-SFT.

\end{document}